\documentclass[letterpaper]{article} 
\usepackage{aaai2026}  
\usepackage{times}  
\usepackage{helvet}  
\usepackage{courier}  
\usepackage[hyphens]{url}  
\usepackage{graphicx} 
\usepackage{array} 
\usepackage{relsize}
\usepackage{bbding}
\usepackage{booktabs}
\usepackage{comment}
\usepackage{graphicx}
\usepackage{adjustbox}
\usepackage{multirow}
\usepackage{booktabs}
\usepackage{amssymb}
\usepackage{array}  

\usepackage[switch]{lineno}

\usepackage{natbib}  
\usepackage{caption} 
\usepackage{algorithm}
\usepackage{algorithmic}
\usepackage{amsmath}
\usepackage{newfloat}
\usepackage{listings}
\DeclareCaptionStyle{ruled}{labelfont=normalfont,labelsep=colon,strut=off} 
\floatstyle{ruled}
\newfloat{listing}{tb}{lst}{}
\floatname{listing}{Listing}
\title{IADGPT: Unified LVLM for Few-Shot Industrial Anomaly Detection, Localization, and Reasoning via In-Context Learning}
\author{
    Mengyang Zhao\textsuperscript{\rm 1},
    Teng Fu\textsuperscript{\rm 1, 2},
    Haiyang Yu\textsuperscript{\rm 1},
    Ke Niu\textsuperscript{\rm 1},
    Bin Li\textsuperscript{\rm 1}
}
\affiliations{
    \textsuperscript{\rm 1}Fudan University \quad
    \textsuperscript{\rm 2}ByteDance Inc.
}

\usepackage{bibentry}

\begin{document}
\nocopyright
\maketitle
\begingroup
\renewcommand\thefootnote{}%
\endgroup
\begin{abstract}


Few-Shot Industrial Anomaly Detection (FS-IAD) has important applications in automating industrial quality inspection. Recently, some FS-IAD methods based on Large Vision-Language Models (LVLMs) have been proposed with some achievements through prompt learning or fine-tuning. However, existing LVLMs focus on general tasks but lack basic industrial knowledge and reasoning capabilities related to FS-IAD, making these methods far from specialized human quality inspectors. To address these challenges, we propose a unified framework, IADGPT, designed to perform FS-IAD in a human-like manner, while also handling associated localization and reasoning tasks, even for diverse and novel industrial products. To this end, we introduce a three-stage progressive training strategy inspired by humans. Specifically, the first two stages gradually guide IADGPT in acquiring fundamental industrial knowledge and discrepancy awareness. In the third stage, we design an in-context learning-based training paradigm, enabling IADGPT to leverage a few-shot image as the exemplars for improved generalization to novel products. In addition, we design a strategy that enables IADGPT to output image-level and pixel-level anomaly scores using the logits output and the attention map, respectively, in conjunction with the language output to accomplish anomaly reasoning. To support our training, we present a new dataset comprising 100K images across 400 diverse industrial product categories with extensive
attribute-level textual annotations. Experiments indicate IADGPT achieves considerable performance gains in anomaly detection and demonstrates competitiveness in anomaly localization and reasoning. We will release our dataset in camera-ready.

\begin{figure}[t]
  \centering
  \includegraphics[width=0.485 \textwidth]{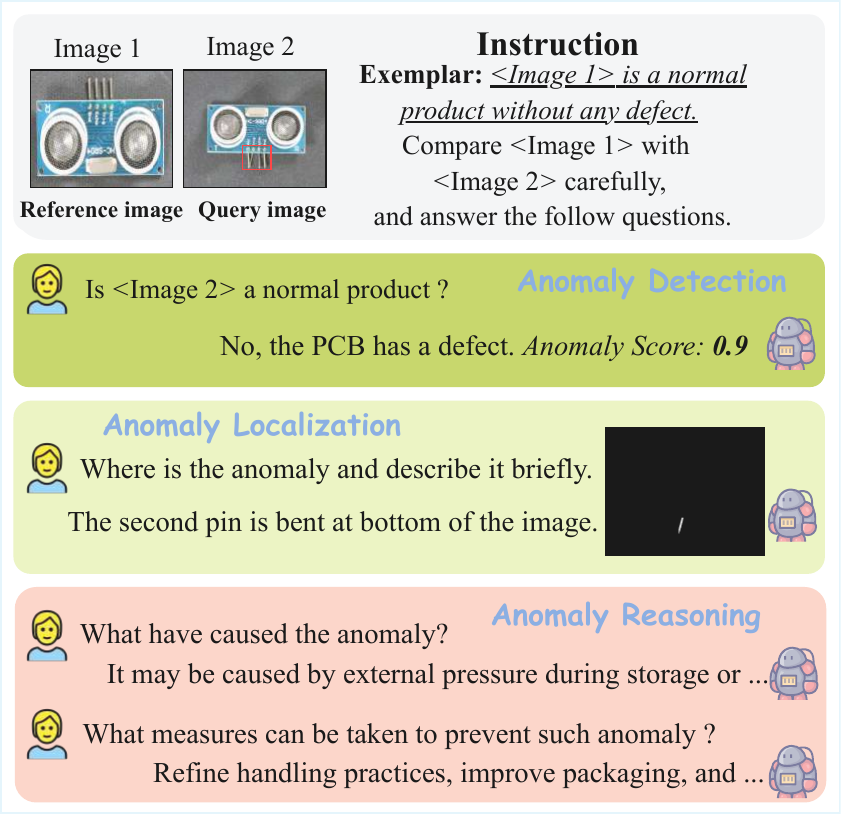}
  \caption{Our model is capable of performing anomaly detection to output an anomaly score, localization to give a pixel mask, and complex reasoning tasks. We treat the reference image as the exemplar context and the query image as the question, enabling one-for-all generalization without updating model parameters for novel classes.}
  \label{fig:intro}
\end{figure}

\end{abstract}

\section{Introduction}
\label{sec:intro}

Industrial Anomaly Detection (IAD) aims to identify the presence of defects in industrial images, and localization refers to pixel-level segmentation of abnormal regions. In addition, anomaly reasoning has also been proposed ~\cite{xu2025towards}, which usually refers to reasoning tasks such as explaining anomalous regions, describing the effects of anomalies, providing possible solutions, etc. Recently, FS-IAD (FS-IAD) methods have been widely investigated since those methods can handle novel industrial products while requiring only a few normal examples. 

Inspired by the excellent zero-/few-shot performance of CLIP ~\cite{clip}, WinCLIP ~\cite{winclip} and PromptAD ~\cite{promptad} design normal/abnormal text prompts and compute the similarity between vision features and normal/abnormal text prompts in the latent space. Despite some progress in zero- or few-shot performance, these methods require redesigning or fine-tuning the prompt for new products, and they are therefore referred to as one-for-one approaches (\textit{i.e.}, training a separate model or prompt for each product). Subsequently, InCTRL ~\cite{Inctrl} and IIPAD ~\cite{oneforall} proposed one-for-all frameworks (\textit{i.e.}, a single model handling all product categories without per-category tuning) using external modules or prompt learning strategy. Despite achieving comparable performance with the one-for-one approach, they remain limited by the CLIP framework, rely on image-text similarity, and are incapable of reasoning due to lack of a Large Language Model (LLM) component.

AnomalyGPT ~\cite{anomalygpt} is the first work to fine-tune an LVLM  for IAD task, which can output anomaly detection and coarse localization results using language. However, it tunes LVLM by adding a learnable embedding as the input of LLM due to a lack of sufficient data, which hardly makes LVLM an IAD expert. 
Anomaly-OV ~\cite{xu2025towards} releases an instruction tuning dataset named Anomaly-Instruct-125k with the detection and reasoning task designed, and fine-tunes an LVLM with a well-designed anomaly expert module under a zero-shot setting.
Some works focus on bootstrapping closed-source LVLM to solve IAD. However, both Cao \textit{et al.} ~\cite{gpt-4v} and Zhang \textit{et al.} ~\cite{gpt-4v-ad} show that even the powerful GPT-4 still has enormous room for improvement in the FS-IAD task.
Despite all the above methods having achieved some progress, they struggle to unleash the power of LVLMs for FS-IAD and are far from the way humans detect anomalies. In addition, existing FS-IAD work does not attempt anomaly reasoning, which ignores the reasoning power of LVLMs and leaves the FS-IAD task lacking in explainability and extensibility.

To this end, we propose a unified model named IADGPT based on open-source LVLM, which can handle diverse novel products no need further tuning, and performs anomaly detection, localization, and reasoning tasks simultaneously, truly realizing \textit{one-for-all}. Specifically, we utilize the logits output by LLM module to compute the image anomaly score, the attention map between visual tokens and some keywords in the prompt text such as \textit{'defect'} and \textit{'anomaly'} to compute the pixel anomaly score, and the language output to perform reasoning, which makes IADGPT no longer require any additional output head. In addition, we propose to treat the few-shot image as exemplar and query image as question, thus allowing IADGPT to naturally implement in-context training and inference without more cumbersome design. Therefore, when facing novel products, IADGPT does not need to optimize learnable prompt embedding or model parameters using the few-shot images.

To adapt the IADGPT to the FS-IAD task, we propose a progressive training strategy containing three stages. Inspired by human quality inspectors, we propose two key competencies critical to FS-IAD: \textit{basic industrial knowledge acquisition} and \textit{discrepancy-aware learning}. On one hand, human experts possess an understanding of fundamental industrial attributes such as structure, material, and surface; on the other hand, they are highly sensitive to visual discrepancies at multiple levels. These capabilities form the foundation of their strong anomaly detection ability. To replicate this abilities, we first design the two training stages to learn fundamental attributes from image-text pairs and to enhance discrepancy awareness through a series of difference perception tasks, respectively. Then, to support these training objectives, we construct a new dataset comprising 100K images across 400 industrial product categories. Compared to existing datasets \cite{visa,Mvtec}, it offers greater category diversity and includes extensive attribute-level textual annotations. As a complement, we also collected almost all existing datasets and constructed them as instruction fine-tuning datasets. Finally, we perform anomaly detection, localization, and reasoning training in the third stage through in-context learning, enabling IADGPT to acquire multi-tasking and one-for-all capabilities.

We evaluate our method at multiple benchmarks, and the experiments indicate IADGPT achieves considerable performance gains in anomaly detection and demonstrates competitiveness in anomaly localization and reasoning. In general, our contributions can be summarized as follows: 
\begin{itemize}
\item We are the first to propose a one-for-all FS-IAD framework trained via in-context learning based on large-scale data, i.e., IADGPT, which is capable of anomaly detection, localization, and reasoning tasks in one go and can handle multiple novel products without further tuning.

\item We design a progressive training strategy inspired by human quality inspectors to gradually guide the IADGPT to acquire basic industrial knowledge, discrepancy-aware ability, which contributes to the model's anomaly detection, localization, and reasoning ability.
\item We propose to utilize in-context learning to handle the FS-IAD task and efficiently utilize the logits, attention map, and language output of LVLM, which yield the unified IADGPT, enabling One Model For All product (i.e., one-for-all) and accomplishing anomaly detection, localization, and reasoning all at once. 
\item We present a large dataset containing 100K images from 400 industrial product categories with extensive attribute-level textual annotations, which makes it rare and contributing in the community.

\end{itemize}

\section{Related Work}
\label{sec:formatting}

\subsection{ Large Vision-Language Models}

CLIP ~\cite{clip} is one of the most commonly used vision-language model, which trained on web-scale image-text and shows strong zero-shot ability. InstructBLIP ~\cite{InstructBLIP} introduces an instruction-aware query transformer based on BLIP2, which can extracts informative features tailored to the given instruction. 
LLaVA ~\cite{liu2023llava} and MiniGPT4 \cite{zhu2023minigpt} introduced visual instruction tuning strategies to enhance instruction following capabilities in LVLMs. Qwen2-VL ~\cite{qwen2-vl} and Qwen2.5-VL ~\cite{Qwen2.5-VL} introduces a naive dynamic resolution mechanism based on Qwen-VL ~\cite{qwenvl}, which enables the model to dynamically process images of varying resolutions into different numbers of visual tokens.  In addition to the open-source LVLM described above, closed-source LVLMs, like GPT-4V ~\cite{gpt4} and GPT-4o, tend to exhibit stronger performance on many general tasks. Alternatively, LVLM can be migrated to specialize in a particular task type. For example, GOT ~\cite{got} migrated LVLM to the OCR domain to cope with different OCR tasks in various scenarios. Our work is also based on the open-source LVLM and aims to develop its  capabilities for FS-IAD.

\subsection{Zero-/FS-IAD Methods}

Recently, zero- and few-shot industrial anomaly detection (IAD) has gained attention due to its ability to rapidly adapt to novel products without extensive retraining, overcoming the generalization limitations of traditional unsupervised methods that rely heavily on large normal datasets.
Inspired by the zero-/few-shot capabilities of CLIP~\cite{clip}, a series of CLIP-based IAD methods~\cite{winclip,anomalyclip,april-gan,promptad} have leveraged image-text similarity to assess anomaly degrees. For example, WinCLIP~\cite{winclip} utilizes normal/abnormal prompts and multi-scale vision-text similarity, while AnomalyCLIP~\cite{anomalyclip} introduces object-agnostic prompt learning. PromptAD~\cite{promptad} combines handcrafted and learnable prompts to maximize the separation between normal and abnormal embeddings. Cao~\textit{et al.}~\cite{gpt-4v} designed tailored prompts for GPT-4V, and GPT-4V-AD\cite{gpt-4v-ad} integrates segmentation model (e.g., SAM~\cite{sam}) with GPT-4V-driven patch-level anomaly scoring. AnomalyGPT~\cite{anomalygpt} is the first to fine-tune an LVLM for anomaly detection and localization using language. Compared to AnomalyGPT, we utilize a larger, more diverse dataset and conduct full-parameter training. 

The work most similar to ours is Anomaly-OV ~\cite{xu2025towards}, which utilizes an additional anomaly expert to optimize the normal/abnormal learnable embedding like CLIP-based methods. In addition, the anomaly expert outputs a token reflecting the anomaly degree as input to LLM, and the LLM mainly focuses on anomaly reasoning. 
However, our method does not require additional learnable embedding and anomaly experts, and enables end-to-end anomaly detection and reasoning using only LVLM. In addition, Anomaly-OV is concerned with the Zero-shot setting and does not enable pixel-level anomaly localization.

\begin{figure}[tb]
  \centering
2026  \includegraphics[width=0.49\textwidth]{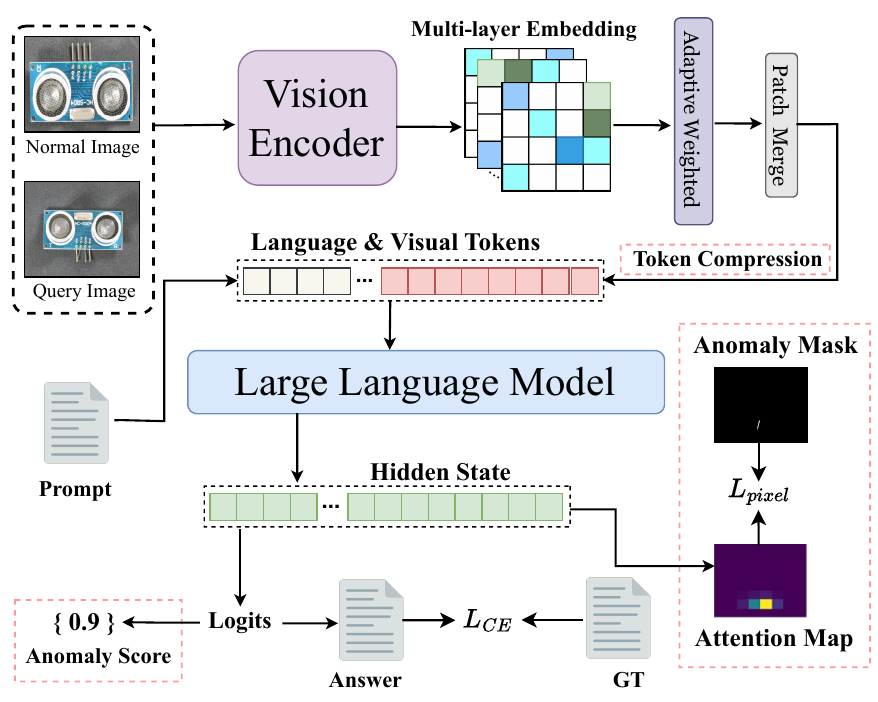}
  \caption{The framework of the proposed model. The components or strategy in the pink rectangular box are only used in the third training stage.
  }
  \label{fig:framework}
\end{figure}

\section{Methodology}
In this section, we first briefly describe the composition of our framework, then present our three-stage training strategy as well as the used data formats, related tasks and optimization objectives, and finally we also briefly introduce the proposed visual token compression module.

\subsection{Framework}

We adapt Qwen2.5-VL-7B~\cite{Qwen2.5-VL} as the basis of our model, mainly consisting of a vision encoder, an LLM, and a patch merge module. As illustrated in Figure \ref{fig:framework}, 
our model contains the basic components of Qwen2.5-VL and adds an adaptive weighted module to fuse the multi-layer visual embedding, which is very useful for IAD \cite{anomalyclip,anomalygpt}. In addition, we design a token compression strategy to avoid the explosion in the number of tokens caused by the few-shot normal image inputs. We design an additional anomaly localization module by supervising the attention map and propose utilizing the logits output by LLM to get the image anomaly score directly. As mentioned before, we also designed a progressive training strategy to migrate our model to specialized FS-IAD tasks.

\subsection{Stage 1: Basic Industrial Knowledge Acquiring}

\begin{figure*}[t]
  \centering
  \includegraphics[width=0.98\textwidth]{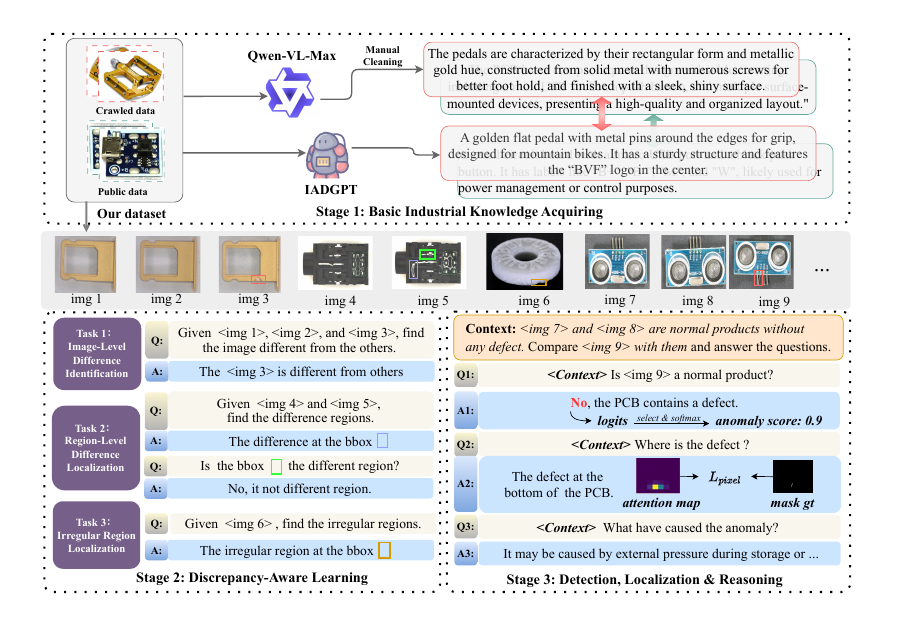}
  \caption{The three-stage training strategy of our IADGPT. The red rectangular boxes are used to label the abnormal regions for demonstration purposes.
  }
  \label{fig:task}
\end{figure*}

To endow IADGPT with basic industrial knowledge akin to that of experienced human inspectors, we design a large-scale training stage focused on attribute-level understanding as shown in Figure~\ref{fig:task}. We construct a dataset of over 1 million image-text pairs describing key properties of industrial products, including shape, material, structural features, and surface quality. While existing IAD datasets (e.g., Real-IAD~\cite{real-iad}, VisA~\cite{visa}, MVTec-AD~\cite{Mvtec}) offer valuable samples, they are limited in diversity and lack descriptive annotations. To address this, we collected approximately 100,000 images covering 400 product categories from the web (via Bing and Baidu), supplemented them with public IAD datasets, and generated rich text descriptions using Qwen-VL-Max, a commercial LVLM with performance comparable to GPT-4. More about the dataset can be viewed in the supplementary materials. The training instance in this stage is represented as an image-text pair $\mathbf{I^{f}_{i}}$:
\begin{equation}
\mathbf{I^{f}_{i}}= \mathbf{(X^{f}_i, C^{f}_i) },
\end{equation} where $\mathbf{X^{f}_i}$ and $\mathbf{C^{f}_i}$ are represent the image and the corresponding description, respectively.
\subsection{Stage 2: Discrepancy-Aware Learning} 
In the FS-IAD task, it is crucial to find the visual difference, potentially leading to anomalies. Therefore, we design the second training stage inspired by human inspectors and formulate three proxy tasks as illustrated in Figure~\ref{fig:task}:

\textbf{Image-Level Difference Identification.}
Given a group of images of the same category, the model is asked to identify the outlier image with real or synthetic defects, enabling coarse-grained discrepancy-aware under image-level supervision.

\textbf{Region-Level Difference Localization.}
The model is asked to compare paired images of the same product type, locate the difference regions, and identify whether some regions are different, enhancing fine-grained discrepancy perception.

\textbf{Irregular Region Localization.}
In the absence of a reference image, the model should localize visual irregularities caused by real or synthetic anomalies from the query image.

These tasks collectively improve IADGPT’s sensitivity to subtle deviations and visual discrepancy, and the final input instance $\mathbf{I^{s}_{j}}$ of the second stage can be denoted as follows:

\begin{equation}
\mathbf{I^{s}_{j}}= \mathbf{(\{P^{s}_1, ... P^{s}_n\}, Q^{s}_j, A^{s}_j) },
\end{equation} where  $\mathbf{\{P^{s}_1, ... P^{s}_n\}}$ denotes the given images, and the $\mathbf{Q^{s}_j}$ and $ \mathbf{A^{s}_j}$ represent the question and answer, respectively.

\subsection{Stage 3: Detection, Localization, and Reasoning}

Existing LVLMs ~\cite{mic,video-llava,LLaVA-NeXT} tend to support more than one image and long text as input at once. Therefore, we can feed the complete instance data of the FS-IAD task to LVLM at once easily, and each instance $\mathbf{I^{t}_{k}}$ can be formed as follows:
\begin{equation}
\mathbf{I^{t}_{k}}= \mathbf{(\{P^{t}_1, ... P^{t}_n\}, X^{t}_k, Q^{t}_k, A^{t}_k) },
\end{equation} where  $\mathbf{\{P^{t}_1, ... P^{t}_n\}}$ denotes the normal example images, $\mathbf{X^{t}_k}$ is the image need to predicted, and the $\mathbf{Q^{t}_k}$ and $ \mathbf{A^{t}_k}$ represent the question and answer, respectively.  Thus, IADGPT can be trained in few-shot context and does not need to utilize the few-shot normal examples to update prompt embedding or model parameters when confronted with novel products.

In the anomaly detection task, we leverage the token-level logits from the LLM's output to compute an image-level anomaly score. As shown in \textit{Q1\&A1} of Figure~\ref{fig:task}, we extract the logits corresponding to the tokens \textit{``Yes''} and \textit{``No''} from the model's prediction and use a softmax operation to obtain a normalized probability distribution. The softmax-normalized probability of \textit{``No''} is used as the image anomaly score $S_{{img}}$, indicating the model’s confidence that the image is defective:
\begin{equation}
S_{{img}} = \frac{e^{z_{\text{no}}}}{e^{z_{\text{yes}}} + e^{z_{\text{no}}}},
\end{equation}

\noindent where \( z_{\text{yes}} \), \( z_{\text{no}} \) denote the logits values corresponding to the tokens \textit{``Yes''} and \textit{``No''}, respectively. In the task, we can use the cross-entropy loss $L_{CE}$ to optimize the whole answer without any other loss function.

In anomaly localization, pixel-level anomaly scores are derived from the attention between visual tokens of the query image and anomaly-related words (e.g.,\textit{``defect''},\textit{``abnormal''}). Attention weights are averaged across multiple layers of the LLM and normalized within the batch to serve as the predicted likelihood of anomalous at the pixel level as the pixel-level anomaly score $S_{\text{pix}}$:

\begin{equation}
S_{{pix}} = \frac{ \left( \frac{1}{L} \sum_{\ell=1}^{L} \mathbf{A}^{(\ell)}_{\text{img} \rightarrow \text{anomaly}} \right) - \mathbf{A}_{min} }{ \mathbf{A}_{max} - \mathbf{A}_{min} + \varepsilon },
\end{equation}
\noindent where \( \mathbf{A}^{(\ell)}_{\text{img} \rightarrow \text{anomaly}} \) is the attention map at layer \( \ell \) from query image tokens to anomaly-related text token;
 \( L \) is the number of attention layers. We use the anomaly mask $S_{gt}$ to supervise  $S_{\text{pix}}$ using the following function:
\begin{equation}
\begin{aligned}
L_{pixel} =\ & \text{Focal}([\hat{S}_{pix}, {S}_{pix}], S_{gt})\\
& + \text{Dice}(\hat{S}_{pix}, I - S_{gt}) + \text{Dice}({S}_{pix}, S_{gt}) ,
\end{aligned}
\end{equation} where $\hat{S}_{pix}$ is the normal score calculated using normal related text token similar to ${S}_{pix}$, where $\text{Focal}$ and $\text{Dice}$ are the focal loss \cite{focal}  and the Dice loss \cite{dice}, respectively.

Regarding the anomaly reasoning task, we utilized the Anomaly-Instruct-125k dataset provided by Anomaly-OV to train our model. This dataset contains a variety of reasoning tasks, such as describing anomalous regions, explaining the causes and effects of anomalies, and possible subsequent measures for improvement, and we show one of them in the \textit{Q3\&A3} of Figure~\ref{fig:task}. In training reasoning tasks, we can utilize the cross-entropy loss directly.

\subsection{Visual Tokens Compression}

Unlike previous models that require only a query image during inference, our model additionally takes few-shot normal examples as input. This introduces a large number of visual tokens, especially with high-resolution images or larger shot counts, leading to increased computational cost and degraded inference efficiency. To this end, we introduce a visual token compression module tailored for the FS-IAD task, built upon the patch merge module.

\begin{table*}[h]
\centering
\newcolumntype{P}[1]{>{\centering\arraybackslash}p{#1}}
\newcolumntype{L}[1]{>{\raggedright\arraybackslash}p{#1}}  
\begin{tabular}{
    P{0.55cm}   
    L{4.8cm}   
    P{1.52cm}  
    P{1.52cm}
    P{1.52cm}
    P{1.52cm}  
    P{1.52cm}
    P{1.52cm}
}
\toprule
\multirow{2}{*}{\centering Setup}
  & \multirow{2}{*}{\makebox[4.8cm][c]{Method}}
  & \multicolumn{3}{c}{MVTec} & \multicolumn{3}{c}{VisA} \\
\cmidrule(lr){3-5} \cmidrule(lr){6-8}
& & 1-shot & 2-shot & 4-shot & 1-shot & 2-shot & 4-shot \\
\midrule
\multirow{7}{*}{\rotatebox{90}{One-for-One}}
&PaDiM \cite{padim} & $76.6,89.3$ & $78.9,91.3$ & $80.4,92.6$ & $62.8,89.9$ & $67.4,92.0$ & $72.8,93.2$ \\
&PatchCore \cite{patchcore} & $83.4,92.0$ & $86.3,93.3$ & $88.8,94.3$ & $79.9,95.4$ & $81.6,96.1$ & $85.3,96.8$ \\
&FastReCon \cite{fastrecon} & ${-,-}$ & $91.0,95.6$ & $94.2,97.0$ & ${-,-}$ & ${-,-}$ & ${-,-}$ \\
&WinCLIP \cite{winclip}& ${93.1,95.2}$ & ${94.4,96.0}$ & ${95.2,96.2}$ & ${83.8,96.4}$ & ${84.6,96.8}$ & ${87.3,97.2}$ \\
&RWDA \cite{rwda}& $93.3,-$ & $94.0,-$ & $94.5,-$ & $83.4,-$ & $85.6,-$ & $86.6,-$ \\
&PromptAD \cite{promptad}& ${\underline{94.6},\underline{95.9}}$ & ${\underline{95.7},\underline{96.2}}$ & ${\underline{96.6},\underline{96.5}}$ & ${\underline{86.9},\underline{96.7}}$ & ${\underline{88.3},\underline{97.1}}$ & ${\underline{89.1},\underline{97.4}}$ \\
&KAG-prompt\cite{kagprompt} & $\mathbf{95.8,96.2}$ & $\mathbf{96.6,96.5}$ & $\mathbf{97.1,96.7}$ & $\mathbf{91.6,97.0}$ & $\mathbf{92.7,97.4}$ & $\mathbf{93.3,97.7}$ \\
\midrule

\multirow{7}{*}{\rotatebox{90}{One-for-All}}
&WinCLIP* \cite{winclip} & ${92.8,92.4}$ & ${92.7,92.4}$ & ${94.0,92.9}$ & ${83.1,94.6}$ & ${83.7,95.1}$ & ${86.1,95.2}$ \\
&AnomalyGPT \cite{anomalygpt}& $94.1,\underline{95.3}$ & $95.5,95.6$ & ${\underline{96.3},96.2}$ & ${\underline{87.4},96.2}$ & ${\underline{88.6},96.4}$ & ${\underline{90.6},96.7}$ \\
&PromptAD* \cite{promptad} & ${86.3,91.8}$ & ${89.2,92.2}$ & ${90.6,92.4}$ & ${80.8,96.3}$ & ${84.3,\underline{96.9}}$ & ${85.7,\underline{97.2}}$ \\
&InCTRL \cite{Inctrl}& ${-,-}$ & ${94.0,-}$ & ${94.5,-}$ & ${-,-}$ & ${85.8,-}$ & ${87.7,-}$ \\
&One-Nor \cite{one-to-normal}& $-,-$ & ${95.1,-}$ & ${95.6,-}$ & ${-,-}$ & ${87.2,-}$ & ${88.6,-}$ \\
& IIPAD \cite{oneforall} & ${\underline{94.2}, \mathbf{96.4}}$ & ${\underline{95.7},\mathbf{96.7}}$ & ${96.1,\mathbf{97.0}}$ & ${85.4,\mathbf{96.9}}$ & ${86.7,\mathbf{97.2}}$ & ${88.3,\mathbf{97.4}}$ \\
\cmidrule(lr){2-8}
&Ours & $\mathbf{96.3,96.4}$ & $\mathbf{96.9},\underline{96.5}$ & $\mathbf{97.3},\underline{96.7}$ & $\mathbf{91.3},\underline{96.7}$ & $\mathbf{92.2},\mathbf{97.2}$ & $\mathbf{93.4,97.4}$ \\
\bottomrule
\end{tabular}
\caption{Each cell reports image-level (left) and pixel-level (right) AUC (in \%). $-$ indicates that relevant data is not reported.  * indicates that the method is applied to one-for-all scenarios reported by IIPAD. The best and second-best results are respectively marked in bold and underlined. }
\label{ad_al}
\end{table*}

Given that normal examples often share high visual similarity, we aim to retain distinctive features while discarding redundant information. Specifically, we use the first example image $\mathbf{P^{t}_1}$ as a reference and compute the cosine similarity between its patch embeddings $\mathbf{p_1}$ and those of each remaining image. The resulting similarity matrix $\mathbf{M^i}$ is defined as:

\begin{equation}
\mathbf{M^i} = \frac{\langle \mathbf{p_1},\mathbf{p_i} \rangle}{\|\mathbf{p_1}\| \|\mathbf{p_i}\|}
\end{equation} Next, we selected the maximum value of the similarity matrix $\mathbf{M^i}$ in the second dimension to get the most similar value of all patches in the current image $\mathbf{P^{t}_i}$ to the first image $\mathbf{P^{t}_1}$:
\begin{equation}
\mathbf{{\hat {M}^i}} = \max_{j=1}^{L_1} \mathbf{M^i_{j,k}} \quad \text{for } k = 1, 2, \ldots, L_i
,\end{equation} where $L_1$ and $L_i$ denote the number of visual tokens corresponding to the first image $\mathbf{P^{t}_1}$ and the current image $\mathbf{P^{t}_i}$, respectively.
Finally, we filter the indexes of top-k most similar patches for each image as the deleted set.

\begin{equation}
U = \mathlarger{\mathlarger{ \mathbf{\cup}}} \underset{i\in [1, n]}{\operatorname*{Index}} (topk(\mathbf{\hat{M}^i}))
\end{equation}


\section{Experiments}
In this section, we introduce the benchmark dataset used in the experiment and the metrics used to evaluate the related methods. In addition, we compare IADGPT with the existing SOTA methods to validate its competitiveness. Finally, we organize ablation experiments to validate the effectiveness of all the proposed modules. For space reasons, some of experiments will be shown in the supplementary material.

\subsection{Datasets and Evaluation Metrics}
We conduct experiments on three datasets: MVTec-AD~\cite{Mvtec}, VisA~\cite{visa}, and the recently proposed Visa-D\&R~\cite{xu2025towards}. MVTec-AD (15 classes) and VisA (12 classes) are widely used in anomaly detection and localization. Following the few-shot setting of AnomalyGPT and PromptAD, we use only their test sets for evaluation, without accessing the training data. Visa-D\&R builds upon VisA by introducing reasoning annotations via GPT-4o and manual verification, requiring LVLMs to infer anomaly causes and suggest corrective measures.

Following prior works~\cite{anomalyclip, promptad}, we use image-level and pixel-level the Area Under the Receiver Operating Characteristic (AUC) to evaluate anomaly detection and localization, respectively. To assess reasoning performance, we adopt Sentence-BERT (SBERT)~\cite{bert} and GPT-Score, as in Anomaly-OV.

\subsection{Implementation Details}
We train IADGPT on 8 Tesla H100 GPUs in three stages. The first stage uses a global batch size of 64 for 3 epochs, while the second and third stages use a batch size of 32 for 1 epoch each. All modules are trained in the first two stages, with the vision encoder frozen in the third. We adopt the AdamW~\cite{adamw} optimizer and a cosine annealing scheduler~\cite{sdrg}, with learning rates of 1e-5, 1e-5, and 2e-6 for the respective stages. To preserve previously learned capabilities, 50\% of the first-stage data is used in the second stage, and 80\% of the second-stage data in the third. After token compression, 20\% of visual tokens $\mathbf{P^{t}_i}$ are retained. For multi-layer visual embedding, we select outputs from layers 8, 16, 24, and 32 of the vision encoder; for attention maps, we use  7, 14, 21, and 28 layers of the LLM, inspired by AnomalyGPT.  

\subsection{Main Results}

To validate the performance of IADGPT in anomaly detection and localization, we compare it with existing methods on MVTec-AD and VisA datasets, as shown in Table \ref{ad_al}. We selected representative one-for-all works such as AnomalyGPT\cite{anomalygpt}, InCTRL\cite{Inctrl},One-Nor\cite{one-to-normal}, MetaUAS\cite{meta-uas}, IIPAD\cite{oneforall}, InInPL\cite{ininpl} as the comparison methods.

Our method achieves a significant improvement in anomaly detection compared to these one-for-all methods, outperforming the IIPAD method by 1.50\% and 5.50\% on average for the three settings on the MvTec and VisA datasets, respectively. However, in anomaly localization, our method achieved a slight performance drop of 0.17\% and 0.07\% on average compared to IIPAD.
This is mainly because the pixel-level AUC has little room for improvement (most methods have exceeded 96\%) and LVLM does not specialize in pixel-level localization. Besides the one-for-all works, we also compare our method with the representative one-for-one works such as PaDiM\cite{padim}, PatchCore \cite{patchcore}, FastRecon \cite{fastrecon}, WinCLIP\cite{winclip}, RWDA\cite{rwda}, PromptAD\cite{promptad}, and KAG-prompt\cite{kagprompt}. Our method achieves almost comparable performance to the SOTA method, KAG-prompt. It is worth noting that the one-for-all setting is more challenging than the one-for-one setting, as discussed in the introduction. As shown in Table~\ref{ad_al}, their performance drops significantly when one-for-one approaches (i.e., methods designed for per-product prompts or models, such as PromptAD* and WinCLIP*) are directly applied to the more challenging one-for-all scenario.
\begin{figure}[t]
  \centering
  \includegraphics[width=0.48 \textwidth]{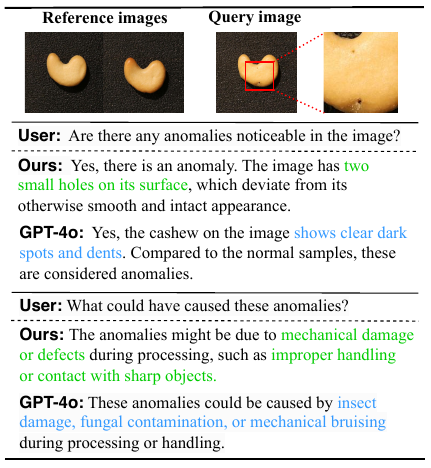}
  \caption{Anomaly image cashew due to small holes. The rightmost image is used to zoom for detail. Fonts with colors are used to highlight key words.}
  \label{fig:vis}
\end{figure}
\begin{table}[t]
\footnotesize
  \centering

  \newcolumntype{P}[1]{>{\centering\arraybackslash}p{#1}}
  \newcolumntype{L}[1]{>{\raggedright\arraybackslash}p{#1}} 

    \begin{tabular}{
    L{2.4cm} 
    P{1.2cm} 
    P{1.2cm} 
    P{1.4cm} 
}
    \toprule
     Model  & Accuracy & SBERT & GPT-Score \\
    \midrule
    GPT-4V*  &0.68 & 0.77 & 5.64    \\
    GPT-4o*  &0.70 & 0.81 & \textbf{6.89}    \\
    Anomaly-OV-7B*    &\underline{0.79} &\underline{0.84} & 6.34    \\
    GPT-4o  &0.76 & 0.83 & 6.85   \\
    Qwen2.5-VL-7B    & 0.54  &  0.69 &  4.80\\
    \midrule
    Ours         & \textbf{0.82} &\textbf{0.86} &\underline{6.51} \\
    \bottomrule
    \end{tabular}%
    \caption{* indicate the performance reported by Anomaly-OV \cite{xu2025towards} with zero-shot setting. Other methods are implemented in 2-shot setting. }
  \label{tab:reasoning}%
\end{table}


To evaluate the proposed method's reasoning ability, we conduct experiments on VisA-D\&R (Table~\ref{tab:reasoning}). Following Anomaly-OV, we report SBERT and GPT-Score to assess semantic consistency between model-generated answers and ground-truth annotations. Unlike Anomaly-OV, we focus exclusively on complex reasoning tasks unrelated to anomaly detection, as detection results are already presented in Table~\ref{ad_al}. Additionally, we report anomaly detection accuracy based on text output, as in Anomaly-OV.
IADGPT consistently outperforms Qwen2.5-VL-7B (without finetuning) across all three metrics and exceeds GPT-4 and GPT-4o in accuracy and SBERT. Although its GPT-Score is slightly lower than GPT-4o, the gap is narrower than that of other methods, likely due to GPT-4o's stronger reasoning ability and its role in data annotation.  As illustrated in Figure~\ref{fig:vis}, our model effectively identifies anomaly causes and demonstrates competitive performance with GPT-4o.

\subsection{Ablation Studies} 
\textbf{Ablation of the proposed components and strategies.}   In order to verify the effectiveness of the proposed strategies and components, we report the image- and pixel-level AUC of the model and its variants on VisA, and the SBERT score on VisA-D\&R, as shown in Table \ref{tab:abl}. Integrating our components can gradually improve overall performance. With the addition of the first stage, the model’s anomaly reasoning performance on VisA-D\&R improves significantly by 5\%, primarily due to enhanced language description capabilities tailored for industrial products. Incorporating the second stage leads to a notable increase in anomaly detection and localization performance, attributed to the model’s acquired discrepancy-aware ability. Furthermore, introducing multi-layer visual embedding and employing the adaptive weight layer for automatically learning important information result in further incremental gains of 0.7\% and 0.3\% in detection and localization, respectively. Furthermore, we can find that after using visual token compression, even if the second reference image retains only 20\% of the visual tokens, the model still maintains almost full performance and outperforms the 1-shot setting by 0.9\% and 0.5\% on image- and pixel-level AUC, respectively.



\begin{table}[t]
\footnotesize
  \centering

  \newcolumntype{P}[1]{>{\centering\arraybackslash}p{#1}}
    \begin{tabular}{
    P{0.5cm} 
    P{0.5cm} 
    P{0.5cm}
    P{0.5cm} 
    P{0.5cm}
    P{1.4cm}
    P{1.5cm}
}
    
    \toprule
    
     S1 & S2  & S3   &\textit{Layer} &\textit{Comp} & VisA & VisA-D\&R \\
    \midrule
    \XSolidBrush  & \XSolidBrush   & \Checkmark  & \XSolidBrush  & \XSolidBrush    & 88.5, 95.1 & 0.81     \\
    \Checkmark  & \XSolidBrush   & \Checkmark   &  \XSolidBrush& \XSolidBrush   & 88.1, 95.3 & 0.86    \\
    \Checkmark      & \Checkmark      & \Checkmark  &  \XSolidBrush& \XSolidBrush   & 91.7, 97.0 & 0.85  \\
    \Checkmark      & \Checkmark      & \Checkmark  & \Checkmark & \XSolidBrush   & 92.4, 97.3 & 0.85  \\
     \Checkmark     & \Checkmark    &  \Checkmark  & \Checkmark  &  \Checkmark     & 92.2, 97.2 & 0.86   \\
    \bottomrule
    \end{tabular}%
  \caption{\textit{Comp} and \textit{Layer} represent the visual token compression strategy and the multi-layer embedding. S1, S2, and S3 represent the three training stage, respectively.  The reported performances are at the 2-shot setting.}
  \label{tab:abl}%
\end{table}


\subsection{Discussion}
\textbf{Why FS-IAD Need LVLM?}
Existing CLIP-based methods ~\cite{anomalyclip,promptad} have been able to utilize vision and language information for FS-IAD, so why do we need to use LVLMs? We aim to provide a preliminary theoretical demonstration that LVLMs have more potential on the FS-IAD task than CLIP-based methods. 

We assume that CLIP and LVLM have the same feature extraction ability for both images and text, and obtained the same embeddings  $\mathbf{M}={\mathbf{I}, \mathbf{X}, \mathbf{T}}$ of the input instance, where $\mathbf{I}=\{\mathbf{I_1}, ... \mathbf{I_n\}}$ denotes the image embedding of the \textit{n} normal example images, $\mathbf{X}$ is the image embedding of the query image, and $\mathbf{T}$ is the related prompt text embedding. CLIP-based methods tend to utilize the relationship between the image $\mathbf{X}$ and the text $\mathbf{T}$ as well as between the image $\mathbf{X}$ and the examples $\mathbf{I}$ to handle FS-IAD, which we can express in the following equation:
\begin{equation}
\mathbf{R_{clip}}=F_{c}(f_{ix}(\mathbf{I},\mathbf{X}),f_{xt}(\mathbf{X},\mathbf{T})),
\end{equation} where $F_{c}$ denotes the decision function of CLIP-based methods, $f_{ix}$ and $f_{xt}$ represent the function to calculate the relationships, and $\mathbf {R_{clip}}$ is the prediction result.
Similarly, we can represent the FS-IAD processing of LVLM as follows:
\begin{equation}
\mathbf {R_{lvlm}}=F_{l}(\mathbf{I}, \mathbf{X}, \mathbf{T}).
\end{equation}
Since $\mathbf{I}, \mathbf{X}, \mathbf{T}$ are concatenated and fed into the LLM simultaneously, the LLM is able to capture the relationship between any two of them as well as the relationship between $\{\mathbf{I_1}, ... \mathbf{I_n\}}$. Thus $\mathbf {R_{lvlm}}$ can also be represented as:
\begin{equation}
\mathbf {R_{lvlm}}=F_{l}(f_{ix}(\mathbf{I},\mathbf{X}),f_{xt}(\mathbf{X},\mathbf{T}),f_{it}(\mathbf{I},\mathbf{T}),f_{ii}(\mathbf{I}), \mathbf{O}),
\end{equation}, where $\mathbf{O}$ represent other information. CLIP-based methods typically employ fixed cosine similarity for $f_{ix}$ and $f_{xt}$, limiting their generalization.  In contrast, LVLMs can learn $f_{ix}$ and $f_{xt}$ autonomously from diverse data, enabling greater generalization. Moreover, LVLMs can leverage additional signals from $f_{it}(\mathbf{I},\mathbf{T})$ and $f_{ii}(\mathbf{I})$, allowing for context-aware tolerance to intra-class variability, \textit{e.g.,} mitigating false anomaly detection when $\mathbf{X}$ deviates from $\mathbf{I}$ in highly diverse settings. These advantages position LVLMs as more promising than CLIP-based methods for FS-IAD.

\section{Conclusion}



In this work, we propose IADGPT, a unified large vision-language model tailored for the FS-IAD task.
IADGPT can simultaneously perform anomaly detection, localization, and reasoning and handle diverse novel products without further tuning. We design a human-inspired, three-stage progressive training strategy to equip the model with the basic industrial knowledge acquisition and coarse and fine-grained discrepancy awareness, and multi-task and in-context learning ability. We also cleverly utilize multiple intermediates of IADGPT to obtain anomaly detection, localization, and reasoning results simultaneously without needing additional branches or output heads. Furthermore, we construct a large-scale industrial dataset containing 100K images across 400 diverse product categories with detailed attribute-level annotations. Extensive experiments demonstrate that IADGPT achieves significant gains in anomaly detection and competitive performance in anomaly localization and reasoning tasks.

\newpage
\bibliography{aaai2026}

\clearpage
\twocolumn[
\begin{center}
    \LARGE\bf Supplementary Materials
\end{center}
\vspace{1.2cm}
]

\section{The Used Dataset}
\subsection{Composition of the Dataset}
Although there are existing many excellent open-source IAD datasets, such as Real-IAD ~\cite{real-iad}, VisA ~\cite{visa}, MVTec-AD ~\cite{Mvtec}, etc., they only contain a small number of types of industrial products (Real-IAD contains the most categories with 30), and they only consist of images without description. Therefore, we crawled about 400 types of industrial products on the Internet through Bing and Baidu, which had a total of about 100,000 images. In addition, we also collected the existing open-source IAD datasets as a supplement to the crawled dataset and then labeled all the datasets. The final used dataset is shown in Table~\ref{tab:unlabeled}. Specifically, we  designed prompts to drive Qwen-VL-Max to describe every image from different perspectives, which is a powerful commercial LVLM comparable to GPT-4o on multiple tasks. We provided three different descriptions for each image, encompassing aspects such as overall shape, material, structural details, and surface quality of the industrial product.
\subsection{Visualization of the Dataset}

\begin{table}[b]
  \centering

  \newcolumntype{P}[1]{>{\centering\arraybackslash}p{#1}}
    \begin{tabular}{
    P{2.3cm}
    P{2.7cm} 
    P{2cm}
}
    \toprule
     Retention Rate & Compression Rate   & VisA  \\
    \midrule
    100\%  &  00.00\%     & 93.5, 97.5   \\    
    40.0\%  &  45.00\%    & 93.5, 97.4   \\
    20.0\%  &   60.00\%   & 93.4, 97.4   \\
    10.0\%  &   67.50\%   & 92.6, 97.2   \\
    5.0\%   &   71.25\%   & 91.7, 96.9   \\
    \bottomrule
    \end{tabular}%
  \caption{Performance of IADGPT at different compression rates with 4-shot setting. Each cell reports image-level (left) and pixel-level (right) AUC (in \%). }
  \label{comp}%
\end{table}

\begin{table}[tb]
\centering

  \setlength{\tabcolsep}{1.6mm}{
     \renewcommand{\arraystretch}{1.23}
    \begin{tabular}{c|cccc}
   
      \hline

      \makebox[0.06\textwidth][c]{Dataset}
           &\makebox[0.05\textwidth][c]{\textit{Images}}  &\makebox[0.05\textwidth][c]{\textit{Class}} & \makebox[0.05\textwidth][c]{\textit{Caption}} & \makebox[0.05\textwidth][c]{\textit{Source}}  \\
      
         \hline

     BTAD & 2,830   &  1    &   No    & Real     \\
     DAGM & 11,500     &  10   &  No   & Synthetic      \\
     Eycandies & 15,500     &   10    &   No  & Synthetic \\        
    
      KolektorSDD  & 399    &  1    &   No    & Real      \\
     KolektorSDD2   & 3,335     &  1   &  No  & Real        \\
     MIAD  & 105,000    &   7   &  No   & Synthetic \\        
    STD  & 3,408     &  1    &   No   & Real \\ 
    Read-IAD  & 126,000    &    25    &   No   & Real        \\
          \hline

     Ours & 100,000     &    400    &   Yes   & Real         \\
               \hline
     Total &  367,972     &    461    &   Yes     & Mixed         \\
      \hline
    \end{tabular}}
    \caption{The dataset used in the first training stage. It is worth noting that the Real-IAD consists of 30 classes totaling about 150,000 images, and we used only 25 classes as the training set and the remaining 5 classes were used as the validation set.}
    \label{tab:unlabeled}
\end{table}

\begin{figure*}[htb]
  \centering
  \includegraphics[width=0.92\textwidth]{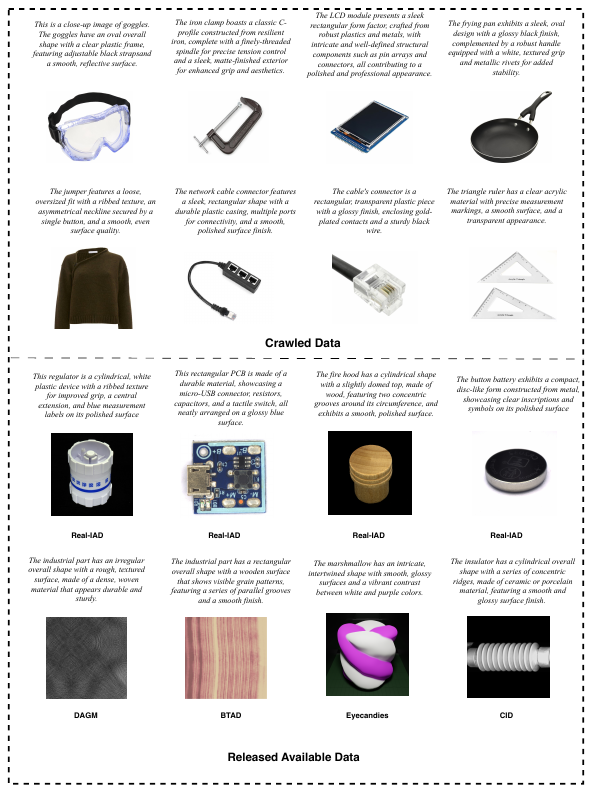}
  \caption{For the released available data, we selected Real-IAD \cite{real-iad}, DAGM \cite{dagm} , Eyecandies \cite{eyecand}, CID \cite{cid}, and BTAD \cite{btad} as the showcase examples.
  }
  \label{fig:dataset}
\end{figure*}

In addition, we present more details of the used dataset. As mentioned in the paper, we constructed a large image-text dataset of more than 1 million by generating descriptions for crawled data and released available data from different perspectives. As shown in Figure ~\ref{fig:dataset}, we can intuitively see that the generated text can clearly describe industrial products from different perspectives, including material and appearance, and so on.

\section{More Implementation Details}
\subsection{Training} In the first stage, we used all the dataset shown in Table~\ref{tab:unlabeled} and the generated image-text pair.
In the second training stage, we mainly use Real-IAD \cite{real-iad}, DAGM\cite{dagm}, KolektorSDD \cite{KolektorSDD}, and KolektorSDD2 \cite{KolektorSDD2} to constructed the data for the designed proxy tasks in the image-text interleaved format. To get more anomalous samples, we were inspired by Cutpaste \cite{cutpaste} to synthesize anomalous data using normal samples. In the end, we obtained about 800, 000 image-text interleaved data. In the third training stage, we designed the anomaly detection and localization under 1-shot, 2-shot, and 4-shot settings. Since the Anomaly-Instruct-125k dataset is zero-shot oriented, in the anomaly reasoning, we constructed some data to few-shot format by utilizing images of the same viewpoint or category as the reference image. However, most data of Anomaly-Instruct-125k were crawled from the web and could not be constructed. In total, about 700, 000 data were constructed for the third training stage.
\subsection{Inference}

When computing image- and pixel-level AUC, we follow previous work \cite{anomalygpt,promptad} for each benchmark dataset by first computing the AUC performance of each category, and then in computing the average AUC of the categories as the final reporting metric. When computing the pixel-level AUC, we treat a category as a large batch and find the maximum value $\mathbf{A}_{max}$ and minimum value $\mathbf{A}_{min}$ in it, and then use Formula 5 in the main text to obtain the normalized pixel-level anomaly score $S_{\text{pix}}$.

\section{Additional Ablation Studies}


\subsection{Visual Token Compression}
To further validate the proposed visual token compression module, we set different compression rates and conducted related experiments as shown in Table~\ref{comp}.  Retention Rate denotes the proportion of unique visual tokens that are preserved from images other than the first reference image. Compression Rate refers to the proportion of visual tokens that are eliminated from the original set of visual tokens corresponding to the 4-shot images. When the Retention Rate is set to 20\%, the model’s performance is nearly unaffected, while 60\% of the visual tokens are discarded. This substantial reduction in token count will significantly improve the inference speed of the model. 
\subsection{Pixel-Level Loss Function}
To verify the effective of pixel-level loss function $L_{pixel}$ in the Formula 6 of main text, we also conducted the ablation experiment as shown in Table~\ref{pixel}. The model with $L_{pixel}$ achieved a huge pixel-level performance gain of 6.2\%, while also delivering a 1.4\% image-level performance gain. This is due to the fact that the pixel-level loss function directs IADGPT to focus on fine-grained anomalous regions, which can provide more accurate supervision signals compared to image-level supervision, which helps both in anomaly detection and localization. 
\subsection{Multi-Layer Attention Weight}
Additionally, to verify the effective of selecting multiple layers of attention weight from LLM (the Formula 5 in the main text), we also conducted the ablation experiment as shown in Table~\ref{pixel}. It can be seen that the multi-layer attention weight is able to provide a decent gain in pixel-level and a slight image-level performance gain of 0.5\% and 0.2\%, respectively. 

\begin{table}[htb]

  \centering

  \newcolumntype{P}[1]{>{\centering\arraybackslash}p{#1}}
    \begin{tabular}{
    P{2.3cm}
    P{2.5cm} 
    P{2cm}
}
    \toprule
     Multi-Layer & $L_{pixel}$   & VisA  \\
    \midrule
   \XSolidBrush   & \XSolidBrush     & 90.6, 90.5   \\    
    \XSolidBrush   & \Checkmark   & 92.0, 96.7   \\
   \Checkmark   & \Checkmark  & 92.2, 97.2   \\
    \bottomrule
    \end{tabular}%
  \caption{Each cell reports image-level (left) and pixel-level (right) AUC (in \%) with 2-shot setting. }
  \label{pixel}%
\end{table}

\section{Qualitative Results}

\subsection{Anomaly Localization}
\begin{figure*}[htbp]
  \centering
  \includegraphics[width=0.999\textwidth]{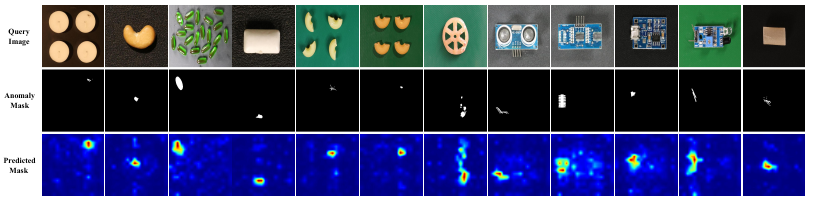}
  \caption{Visualization results for anomaly localization with 1-shot setting on VisA dataset.}
  \label{vis-loc}
\end{figure*}

To further verify the effectiveness of IADGPT for anomaly localization, we conducted visualization experiments as shown in Figure~\ref{vis-loc}. We utilize heat maps to annotate anomalous regions, where deeper colors indicate higher predicted anomaly scores. It can be observed that IADGPT is capable of accurately localizing anomalous areas across a variety of different products.

\subsection{Anomaly Reasoning}
\begin{figure}[htbp]
  \centering
  \includegraphics[width=0.45\textwidth]{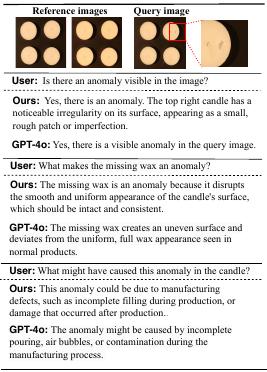}
  \caption{Anomaly reasoning about a chunk of wax missing.}
  \label{qa1}
\end{figure}

\begin{figure}[htbp]
  \centering
  \includegraphics[width=0.47\textwidth]{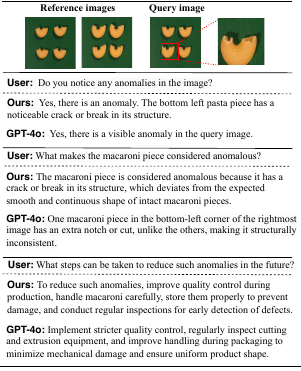}
  \caption{Anomaly reasoning about chipped corners.}
  \label{qa2}
\end{figure}

To further validate the effectiveness of IADGPT for anomaly reasoning, we additionally report some reasoning results as shown in Figure~\ref{qa1} and ~\ref{qa2}.The comparisons of the reasoning processes between GPT-4o and IADGPT reveal that IADGPT can not only provide a coarse localization of anomalies using language but also recognize the manifestations or categories of anomalies. The reasoning outputs are comparable to those generated by GPT-4o. 

\section{Additional Experiment Results}
In this section, we report the performance of IADGPT on the VisA and MvTec benchmarks for each category. We report the image-level and pixel-level AUC performance of IADGPT on MvTec and on the VisA dataset in Tables ~\ref{mvtec} and ~\ref{visa}, respectively. This directly validates the capability of IADGPT for anomaly detection and localization across various novel products. Experimental results indicate that IADGPT consistently demonstrates good performance in detecting and localizing anomalies in a wide range of novel products.


\begin{table*}[ht]
\centering
\newcolumntype{P}[1]{>{\centering\arraybackslash}p{#1}}
\newcolumntype{L}[1]{>{\raggedright\arraybackslash}p{#1}}

\begin{tabular}{
    L{2cm}   
    P{1.6cm} 
    P{1.6cm} 
    P{1.6cm} 
    P{1.6cm} 
    P{1.6cm} 
    P{1.6cm} 
}
\toprule
\multirow{2}{*}{Class} & \multicolumn{3}{c}{Image-Level} & \multicolumn{3}{c}{Pixel-Level} \\
\cmidrule(lr){2-4} \cmidrule(lr){5-7}
& 1-shot & 2-shot & 4-shot & 1-shot & 2-shot & 4-shot \\
\midrule
Bottle      & 98.8 & 99.1 & 99.3 & 99.0 & 98.9 & 99.0 \\
Cable       & 92.3 & 93.6 & 94.1 & 97.1 & 97.2 & 97.4 \\
Capsule     & 95.2 & 95.7 & 96.2 & 97.9 & 98.0 & 98.3 \\
Carpet      & 97.6 & 97.9 & 98.3 & 97.4 & 97.6 & 98.1 \\
Grid        & 96.5 & 97.2 & 97.9 & 96.5 & 97.1 & 97.2 \\
Hazelnut    & 98.0 & 98.4 & 98.5 & 97.8 & 97.7 & 97.8 \\
Leather     & 97.5 & 98.3 & 98.4 & 96.2 & 96.5 & 97.0 \\
Metal nut   & 98.0 & 98.7 & 99.0 & 95.6 & 96.0 & 96.1 \\
Pill        & 95.7 & 96.2 & 96.6 & 97.5 & 97.2 & 97.2 \\
Screw       & 91.5 & 92.3 & 92.8 & 95.3 & 95.8 & 96.1 \\
Tile        & 98.3 & 98.8 & 99.4 & 95.7 & 96.2 & 96.2 \\
Toothbrush  & 97.2 & 97.7 & 98.3 & 94.2 & 94.0 & 94.1 \\
Transistor  & 94.1 & 94.9 & 95.5 & 96.0 & 96.3 & 96.2 \\
Wood        & 97.2 & 97.5 & 97.6 & 93.1 & 93.4 & 94.0 \\
Zipper      & 97.0 & 97.4 & 97.8 & 96.3 & 96.3 & 96.5 \\
\midrule
Mean        & 96.3 & 96.9 & 97.3 & 96.4 & 96.5 & 96.7 \\
\bottomrule
\end{tabular}
\caption{Performance (AUC \%) of IADGPT on each category on the MvTec dataset.}
\label{mvtec}
\end{table*}

\begin{table*}[htb]
\centering
\newcolumntype{P}[1]{>{\centering\arraybackslash}p{#1}}
\begin{tabular}{
    p{2cm}   
    P{1.6cm} 
    P{1.6cm} 
    P{1.6cm} 
    P{1.6cm} 
    P{1.6cm} 
    P{1.6cm} 
}
\toprule
\multirow{2}{*}{Class} & \multicolumn{3}{c}{Image-Level} & \multicolumn{3}{c}{Pixel-Level} \\
\cmidrule(lr){2-4} \cmidrule(lr){5-7}
& 1-shot & 2-shot & 4-shot & 1-shot & 2-shot & 4-shot \\
\midrule
Candle     & 94.6 & 95.0 & 95.0 & 98.6 & 98.6 & 98.7 \\
Capsules   & 91.0 & 92.2 & 91.4 & 98.3 & 98.3 & 98.1 \\
Cashew     & 91.4 & 92.0 & 94.5 & 95.5 & 97.0 & 97.3 \\
Chewinggum & 98.0 & 98.3 & 99.3 & 99.2 & 99.5 & 99.6 \\
Fryum      & 88.6 & 89.2 & 92.0 & 95.5 & 95.8 & 96.1 \\
Macaroni1  & 90.1 & 90.4 & 92.8 & 98.9 & 99.0 & 99.2 \\
Macaroni2  & 81.9 & 84.5 & 86.8 & 96.6 & 96.8 & 96.9 \\
PCB1       & 94.2 & 94.0 & 95.3 & 97.5 & 98.3 & 98.4 \\
PCB2       & 89.4 & 90.5 & 90.9 & 95.1 & 95.7 & 95.9 \\
PCB3       & 80.5 & 83.7 & 85.5 & 92.7 & 93.6 & 93.8   \\
PCB4       & 97.1 & 97.4 & 97.7 & 96.3 & 96.4 & 97.0 \\
Pipe fryum &99.1 & 99.3 & 99.6  & 96.8 & 97.4 & 98.1 \\
\midrule
Mean & 91.3 & 92.2 & 93.4 & 96.7 & 97.2 & 97.4 \\
\bottomrule
\end{tabular}
\caption{Performance (AUC \%) of IADGPT on each category on the VisA dataset.}
\label{visa}
\end{table*}

\newpage

\end{document}